%% file: root.tex
\documentclass[letterpaper, 10 pt, conference]{ieeeconf}  %

\IEEEoverridecommandlockouts                              %

\title{\LARGE \bf
Ask-to-Clarify: Resolving Instruction Ambiguity\\through Multi-turn Dialogue
}

\def\onedot{.}
\def\eg{\emph{e.g}\onedot}
\def\ie{\emph{i.e}\onedot}

\newcommand\framework{Ask-to-Clarify}

\author{
Xingyao Lin$^{1,2}$,
Xinghao Zhu$^{3}$,
Tianyi Lu$^{1,2}$,
Guojin Zhong$^{1}$,
Sicheng Xie$^{1,2}$,
Hui Zhang$^{1}$,\\
Xipeng Qiu$^{1,2}$,
Zuxuan Wu$^{1,2}$\textsuperscript{\textdagger},
Yu-Gang Jiang$^{1}$
\thanks{This work is supported by the National Natural Science Foundation of China (Grant No. 62521004) and  the Science and Technology Commission of Shanghai Municipality (No. 24511103100).}
\thanks{\textdagger \ denotes the corresponding author.}
\thanks{$^{1}$College of Computer Science and Artificial Intelligence, Fudan University, Shanghai, China.
}
\thanks{$^{2}$Shanghai Innovation Institute, Shanghai, China.
}
\thanks{$^{3}$Mechanical Systems Control Lab, UC Berkeley, California, USA.
}
}

\usepackage{amsmath} 
\usepackage{amssymb}

\usepackage{graphicx}

\usepackage{caption}
\usepackage{multirow}
\usepackage{array}
\newcolumntype{P}[1]{>{\centering\arraybackslash}p{#1}}
\usepackage{bbding}
\usepackage{booktabs}
\usepackage{arydshln}
\usepackage{makecell}

\usepackage{xcolor}
\definecolor{myblue}{rgb}{0.21, 0.49, 0.74}
\usepackage[colorlinks=true,
            pagebackref=true,
            linkcolor=myblue,
            citecolor=myblue,
            urlcolor=myblue]{hyperref}
\makeatletter
\let\NAT@parse\undefined
\makeatother
\usepackage[numbers,sort&compress]{natbib}
\usepackage{cleveref}

\begin{document}

\maketitle
\thispagestyle{empty}
\pagestyle{empty}

\begin{abstract}
\input{tex/0_abstract}
\end{abstract}
\section{Introduction}
\input{tex/1_introduction}
\section{Related Work}
\input{tex/2_related_work}
\section{Method\label{sec:method}}
\input{tex/3_method}
\section{Experiments}
\input{tex/4_experiment}
\section{Conclusion}
\input{tex/5_conclusion}
{\small
\bibliographystyle{IEEEtran}
\bibliography{references}
}

\end{document}

%% file: tex/0_abstract.tex
Embodied agents are intelligent systems designed to perceive, reason, and act within the physical world. 
While the robotics community has long strived to build such versatile agents, a fundamental limitation persists: most current VLA-based models operate under a rigid ``Listen-and-Act'' paradigm. 
These systems assume instructions are unambiguous and execute them in a passive fashion, preventing them from resolving uncertainty through dialogue. 
To address this, we propose \framework{}, a unified end-to-end framework that seamlessly integrates multi-turn disambiguation dialogue with low-level visuomotor control, eliminating the reliance on high-level action primitives or external planners. 
Specifically, \framework{} synergizes a VLM-based Cognitive Planner with a Diffusion-based Motor Executor. 
To bridge the disparity between high-level disambiguation and low-level execution, we introduce a Semantic-Visual Alignment Adapter, which functions as a cross-modal interface to synthesize semantic intent with visual perceptual streams. 
Furthermore, we observe severe catastrophic forgetting: visuomotor fine-tuning completely erases dialogue capabilities. To overcome this, we propose a two-stage knowledge-insulation training strategy, effectively decoupling dialogue logic from physical manipulation.
Extensive evaluations across 11 real-world tasks demonstrate that \framework{} significantly outperforms existing methods, offering a promising path toward building truly collaborative embodied agents.

%% file: tex/1_introduction.tex
Embodied agents are intelligent systems designed to perceive, reason, and act within the physical world. 
Unlike systems that only solve abstract problems in a virtual space, embodied agents must navigate the complexity and unpredictability of real-world environments.
Beyond these physical challenges, embodied agents face an extra challenge not present in virtual environments~\textemdash~collaboration with humans.
\textit{Collaboration behavior} refers to the agent’s ability to communicate, coordinate, and adapt its actions based on human feedback.
Thus, the ultimate goal of embodied agents is to create collaborators that can actively interact with humans in the real-world environment, not mere executors that passively follow instructions.
\input{figure/teaser}

For decades, the robotics community has strived to build such versatile agents~\cite{kunze2018artificial, duan2022survey}. 
Recently, the advent of Vision-Language Models (VLMs) has catalyzed a paradigm shift, leading to the development of Vision-Language-Action (VLA) models. 
By building on the vast, internet-scale knowledge of VLMs, VLAs have demonstrated remarkable potential for generalization. 
However, a fundamental limitation persists: most current VLA-based agents operate under a rigid ``Listen-and-Act'' paradigm. 
They assume instructions are unambiguous and execute them passively. 
Consider a household scenario where a user commands, ``Give me the mug,'' while multiple mugs sit on a table. 
A standard agent, lacking disambiguation capabilities, would likely hallucinate a target or fail entirely. 
This inability to resolve ambiguity through dialogue prevents current systems from being reliable collaborators in real-world settings.

To address this, prior works such as DialFRED~\cite{dialfred}, TEACh~\cite{teach}, and ASK-TO-ACT~\cite{ask-to-act} have explored interactive disambiguation. 
However, these approaches rely on high-level action primitives (\eg,~\texttt{MoveRight}), restricting them to the simplest manipulation tasks.
On the other hand, state-of-the-art visuomotor policies, such as recent VLAs~\cite{openvla, openvla-oft, pi-0, pi-0.5}, can perform complex manipulation but lack the interactive dialogue capabilities to resolve semantic ambiguity.

In this paper, we bridge this gap by proposing \framework{}, a hierarchical framework that unifies high-level cognitive capabilities with low-level visuomotor control. 
Unlike passive systems, \framework{} functions as an active collaborator: it detects ambiguity, initiates clarification dialogues, and translates the refined intent into precise motor trajectories. 
\Cref{fig:teaser} illustrates this shift: while a passive executor fails under uncertainty, our collaborator actively engages the user to deduce the correct intent before execution.

The design principle of Ask-to-Clarify is to decouple disambiguation from execution while aligning them through a shared
representational interface.
Specifically, \framework{} synergizes a VLM-based Cognitive Planner with a Diffusion-based Motor Executor. 
To bridge the representation gap between high-level disambiguation and low-level execution, we introduce a Semantic-Visual Alignment Adapter. 
This adapter functions as a cross-modal interface, synthesizing the planner's semantic intent with visual perceptual streams.
By modulating instruction embeddings via visual latent states, it produces visually-grounded conditions that precisely guide the executor.
Furthermore, we observe severe catastrophic forgetting when fine-tuning the planner for visuomotor control. Conversely, strictly freezing the planner avoids this erasure but hampers control precision due to feature misalignment. 
To resolve this, we devise a two-stage knowledge-insulation training strategy. 
This strategy first cultivates active disambiguation capabilities, and then exclusively optimizes the adapter and executor while keeping the cognitive parameters frozen. 
This design effectively aligns cognitive capabilities with visuomotor control requirements without eroding the planner's previously acquired knowledge. 
During deployment, a Disambiguation Gate autonomously governs the transition between ``Disambiguation Mode'' and ``Execution Mode'' based on emergent signal tokens.

We evaluate \framework{} across 11 real-world tasks, comparing it against existing state-of-the-art approaches. 
Our framework significantly outperforms these baselines, demonstrating that decoupling disambiguation from execution, while aligning them through a specialized adapter, is key to robust real-world collaboration.

In summary, our contributions are threefold:
\begin{itemize}
  \item We present a novel end-to-end framework that unifies multi-turn disambiguation dialogue with low-level visuomotor control, eliminating the reliance on high-level action primitives or external planners.
  
  \item Through systematic ablation, we observe severe catastrophic forgetting: dialogue capabilities are completely erased by even partial fine-tuning during visuomotor learning. To overcome this, we propose a two-stage knowledge-insulation training strategy. This strategy effectively decouples cognitive capabilities from visuomotor control, ensuring the agent retains its dialogue logic while mastering precise manipulation.

  \item We validate \framework{} through extensive real-world experiments, demonstrating state-of-the-art performance. Furthermore, we show that our proposed Semantic-Visual Alignment Adapter effectively bridges the representation gap between abstract dialogue logic and concrete real-world manipulation, ensuring robust physical collaboration.
\end{itemize}

%% file: figure/teaser.tex
\begin{figure}[ht]
\centering
\includegraphics[width=.99\linewidth]{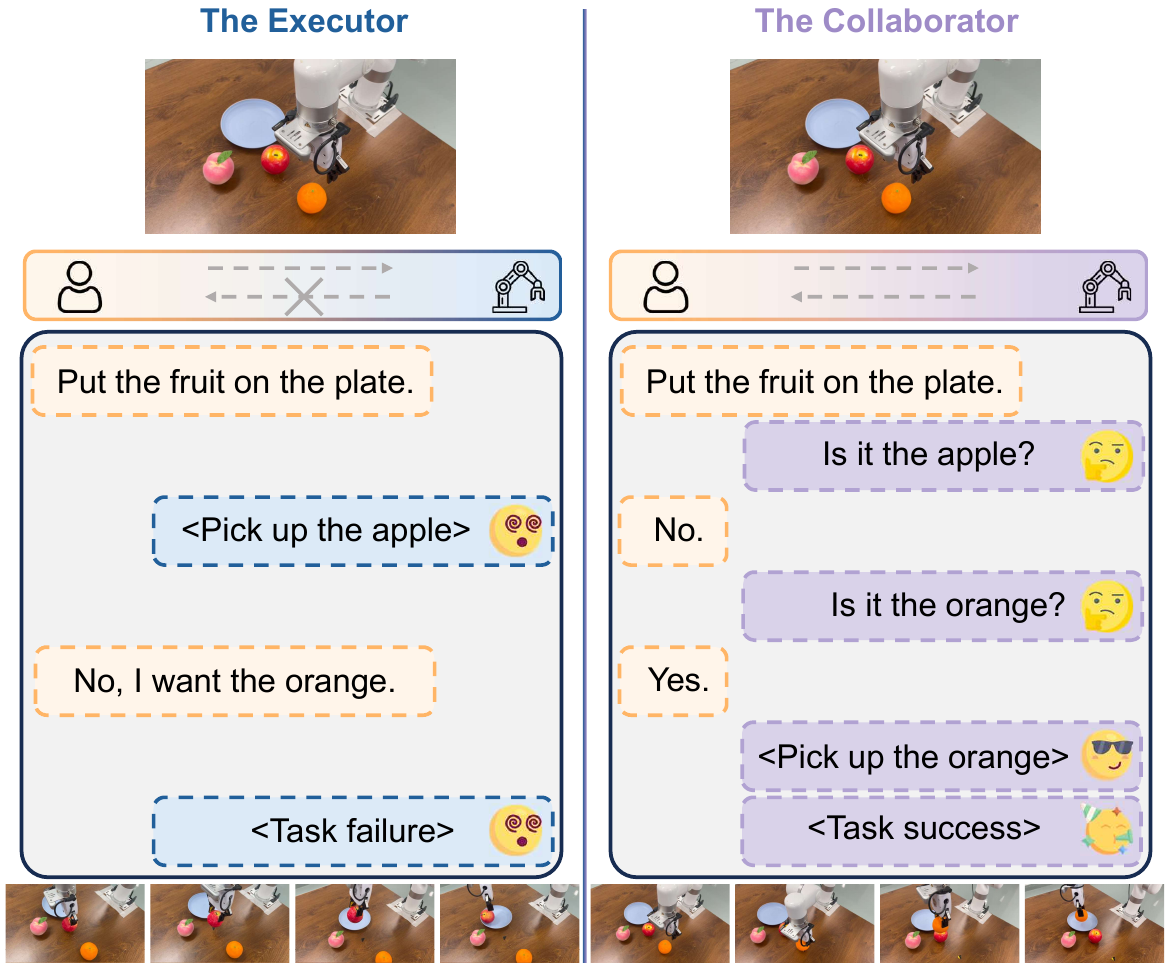} 
\caption{
\textbf{Executor vs. Collaborator.}
Unlike passive executors that fail on ambiguous instructions, active collaborators resolve ambiguity via dialogue.
}
\label{fig:teaser}
\end{figure}

%% file: tex/2_related_work.tex
\subsection{Embodied Agents for Manipulation}
Recent Vision-Language-Action (VLA) models leverage large-scale pretraining to map observations to actions~\cite{openvla, openvla-oft, pi-0, pi0-fast, activemimic,thinkingvla, pi-0.5}. 
One line of work focuses on autoregressive VLAs that sequentially predict action tokens. For instance, OpenVLA~\cite{openvla} maps continuous robot actions to discrete tokens using the VLM's original tokenizer. However, this autoregressive generation is often too slow for high-frequency robot control.

To address this, other works decouple action generation from the VLM using a specialized action expert, enabling faster response times. These methods generally adopt either a hierarchical or a parallel architecture. 
Hierarchical models separate cognition from action generation serially. For example, DexVLA~\cite{dexvla} transmits VLM features to a diffusion-based action expert trained via an embodiment curriculum. 
Conversely, parallel models operate the VLM and the action expert simultaneously. For instance, $\pi_{0}$~\cite{pi-0} combines a pretrained VLM backbone with an independent flow-matching action expert to handle robot-specific inputs.

In this work, we adopt a hierarchical architecture to synergize the VLM's cognitive capabilities with diffusion-based visuomotor control. Crucially, we introduce a Semantic-Visual Alignment Adapter to bridge the representation gap, ensuring high-level intent is effectively translated into precise low-level manipulation.

\subsection{Agents that Collaborate with Humans}
Robots operating in human spaces require natural language interaction~\cite{dialfred, teach, ask-to-act, asking}. Recent methods allow agents to actively ask questions, using the user's responses to better complete tasks. For example, ASK-TO-ACT~\cite{ask-to-act} fine-tunes a VLM to resolve ambiguity by asking clarification questions and controls the robot using oracle high-level actions.
However, most existing approaches rely on high-level action primitives~\cite{ask-to-act, dialfred, teach}. This reliance is insufficient for complex tasks requiring precise manipulation. In contrast, \framework{} advances the field by directly targeting end-to-end low-level control. To achieve this without suffering from severe catastrophic forgetting during visuomotor fine-tuning, we propose a two-stage knowledge-insulation training strategy,
which effectively decouples cognitive capabilities from visuomotor control, ensuring the agent retains its dialogue logic while mastering fine-grained manipulation skills.

%% file: tex/3_method.tex
\input{figure/train}
In this section, we present \framework{}, a collaborative embodied framework designed to unify high-level cognition capabilities with low-level visuomotor control.
Unlike traditional agents that operate in a passive manner~\cite{openvla, rt-2}, \framework{} establishes a bi-directional feedback loop with human users.
The architecture synergizes two core components: a Cognitive Planner, initialized from a VLM~\cite{qwen2-vl}, and a Motor Executor, derived from a pre-trained diffusion policy~\cite{scaledp}.
To bridge the gap between high-level disambiguation and low-level execution, we introduce the Semantic-Visual Alignment Adapter. 
This module achieves a visually-grounded refinement of user intent by modulating linguistic embeddings with visual features via feature-wise affine transformations.
Additionally, we employ a two-stage knowledge-insulation training strategy to ensure that learning fine-grained visuomotor control does not compromise the agent's cognitive capabilities.
We first formally define the collaborative embodied task in \Cref{subsec:task-define}. 
Subsequently, we detail our training methodology in \Cref{subsec:training-strategy}. 
Finally, we describe the inference pipeline in \Cref{subsec:inference}.

\subsection{Task Definition}
\label{subsec:task-define}
We formalize the task of collaborative embodied manipulation, which extends standard instruction following by incorporating an interactive disambiguation phase.
Consider an environment where an agent receives an initial natural language instruction $\mathbf{I}_\mathrm{init}$ alongside an observation $\mathbf{O}$.
In real-world unstructured environments, $\mathbf{I}_\mathrm{init}$ is often ambiguous due to the presence of multiple candidate objects or underspecified constraints (e.g., ``Give me the mug'' when multiple mugs exist).
Existing VLAs typically fail in such scenarios due to their rigid ``Listen-and-Act'' paradigm.

Our task requires the agent to dynamically switch between two modes: \textit{Disambiguation} and \textit{Execution}.
Formally, given $\mathbf{I}_\mathrm{init}$ and $\mathbf{O}$, the agent must determine if ambiguity exists.
If ambiguous, the agent initiates a multi-turn dialogue, generating a clarification query $\mathbf{Q}_i$ at turn $i$ and receiving a user response $\mathbf{A}_i$.
This process iterates until the agent deduces a refined, unambiguous instruction $\mathbf{I}_\mathrm{refined}$.
Mathematically, the resolved intent is derived as $\mathbf{I}_\mathrm{refined} = f_{\mathrm{Disambiguation}}(\mathbf{I}_\mathrm{init}, \mathbf{O}, \mathcal{H})$, where $\mathcal{H} = [(\mathbf{Q}_1, \mathbf{A}_1), \dots, (\mathbf{Q}_n, \mathbf{A}_n)]$ represents the dialogue history.
Upon resolution, the agent transitions to the execution phase, generating a series of action chunks $\mathbf{a}_{t:t+k} = f_{\mathrm{Execution}}(\mathbf{I}_\mathrm{refined}, \mathbf{O})$ to complete the task.

\subsection{Two-stage Knowledge-insulation Training Strategy}
\label{subsec:training-strategy}
Training a unified model to handle both complex cognition and fine-grained motor control often leads to conflicting gradients.
Crucially, we observe severe catastrophic forgetting: dialogue capabilities acquired initially are rapidly erased by even partial fine-tuning during the action generation stage.
To mitigate this, we propose a two-stage knowledge-insulation training strategy.
This paradigm decouples the acquisition of cognitive capabilities from visuomotor skills, ensuring that the agent retains its dialogue logic while mastering physical manipulation.

\textbf{Active Disambiguation}
The primary objective of the first stage is to empower the Cognitive Planner with the ability to detect ambiguity and engage in clarification dialogues.
To construct a robust dataset for this purpose, we employ a scalable data synthesis pipeline.
We first collect images of multiple objects that differ from each other in some but not all attributes,~\eg, two blocks that differ only in color.
Then we use off-the-shelf VLMs to generate the ambiguous instruction, the question-answer pairs, and the correct instruction based on the collected images.
Finally, we use this generated content to construct our dialogue data.
This process yields a diverse ambiguity-solving dialogue dataset, which serves as the training corpus for the planner.

To facilitate mode switching, we introduce signal tokens into the VLM vocabulary: \texttt{<AMBG>} denotes ambiguity, prompting the generation of a clarification question; \texttt{<NOT\_AMBG>} marks the deduction of a resolved instruction; and \texttt{<ACT>} or \texttt{<REJ>} indicates the readiness to execute or a refusal based on feasibility.
By optimizing the planner to predict these tokens, \framework{} learns to autonomously govern its operational mode.
During this stage, we decouple the executor, freezing the vision encoder and fine-tuning only the LLM parameters of our planner. 
This stage efficiently adapts the planner to the collaborative embodied agent while retaining the generalizable knowledge of the VLM.

\textbf{End-to-end Action Generation}
In the second stage, we focus on grounding semantic intent into precise motor actions. 
To bridge the Cognitive Planner and the Motor Executor, we introduce the Semantic-Visual Alignment Adapter which functions as a cross-modal interface.
Specifically, it extracts the refined instruction \(\mathbf{I}_\mathrm{refined}\) and visual features from the planner, synthesizing them into a visually-grounded representation via affine transformations. This ensures that the executor receives a robust, instruction-aware condition. 

Furthermore, to optimize this architecture without compromising existing capabilities, we must address the risk of catastrophic forgetting. 
A monolithic approach of fine-tuning the entire model inevitably erases the planner's previously acquired dialogue logic. 
Consequently, we employ a knowledge-insulation mechanism by freezing the parameters of the Cognitive Planner. This strategy safely isolates the planner's learned knowledge, thereby restricting optimization exclusively to the adapter and executor.
By training on expert demonstrations labeled with correct instructions, the executor learns to generate high-frequency motor trajectories. 
Ultimately, this effectively insulates the high-level cognitive capabilities from the low-level visuomotor control, yielding a collaborative embodied agent that is both an articulate communicator and a dexterous manipulator. 
The overall training pipeline is illustrated in~\Cref{fig:train}.

\subsection{Disambiguation Gate}
\label{subsec:inference}

\input{figure/infer}
During inference, \framework{} operates as an active collaborator, dynamically switching behaviors via our training-free Disambiguation Gate. This gate functions as a router, interpreting the signal tokens emitted by the Cognitive Planner to direct the system's flow.

Specifically, the process initiates with the planner analyzing the initial instruction $\mathbf{I}_\mathrm{init}$ and observation $\mathbf{O}$. 
Upon detecting semantic uncertainty, the planner emits the \texttt{<AMBG>} token, which triggers a suspension of the motor control loop. 
The system then outputs a clarification query to the user, and the subsequent response is appended to the dialogue history for context updating in the next inference cycle.
Conversely, when the intent is clear or successfully disambiguated, the planner generates the refined instruction $\mathbf{I}_\mathrm{refined}$ followed by the \texttt{<NOT\_AMBG>} token. 
This refined intent is then verified; if the target object is present and actionable, the planner releases the \texttt{<ACT>} token. 
This signal activates the Semantic-Visual Alignment Adapter and Motor Executor, translating the refined intent into physical motion. 
In cases of unresolvable conflicts or safety risks (e.g., target absence), the \texttt{<REJ>} token halts the operation immediately. 
This hierarchical inference logic ensures that \framework{} only acts when intent is aligned with reality, effectively solving the ``Listen-and-Act'' problem.
The working flow of the Disambiguation Gate can be found in~\Cref{fig:infer}.

%% file: figure/train.tex
\begin{figure*}[htb]
\centering
\includegraphics[width=.99\textwidth]{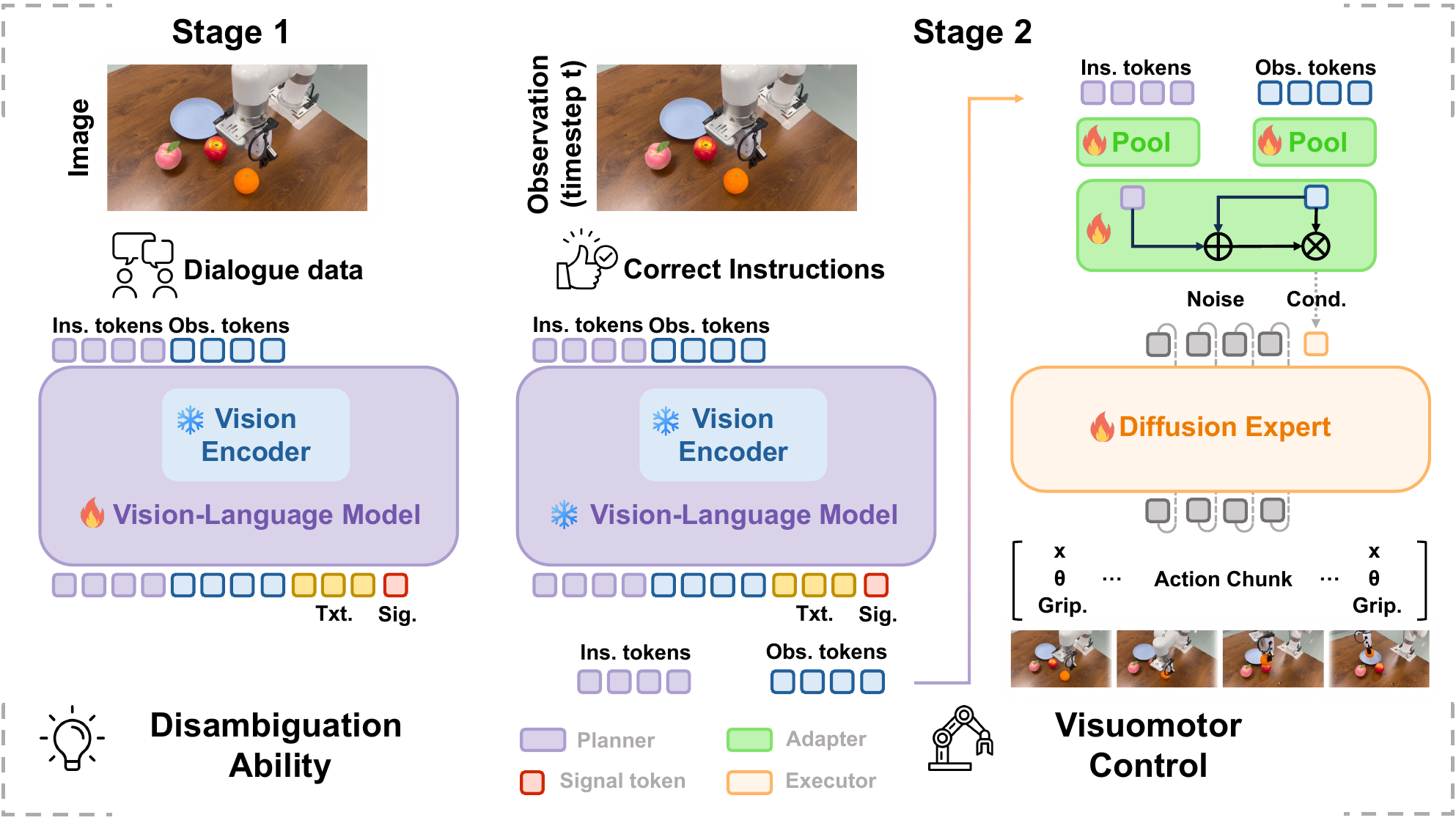} 
\caption{
\textbf{The~\framework{} framework.}
  Stage 1 fine-tunes the planner on dialogue data to acquire disambiguation ability. 
  Stage 2 implements visuomotor control by freezing the planner and optimizing the adapter and executor for action generation. 
}

\label{fig:train}
\end{figure*}

%% file: figure/infer.tex
\begin{figure}[htb]
\centering
\includegraphics[width=.70\linewidth]{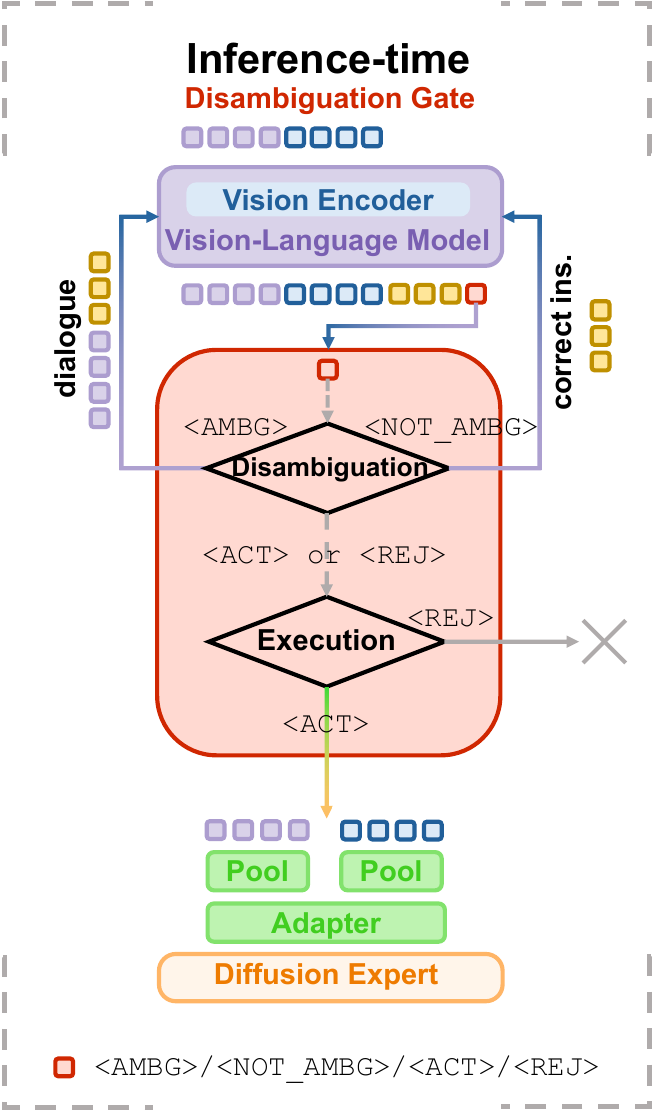} 
\caption{
  \textbf{The Disambiguation Gate.} 
}
\label{fig:infer}
\vspace{-0.7em}
\end{figure}

%% file: tex/4_experiment.tex
\input{figure/task}
The~\framework{} framework is designed to act as a collaborative embodied agent.
To evaluate our framework, we conduct a series of real-world experiments.
First, we assess whether our framework acquires collaborative embodied abilities and evaluate its overall performance against state-of-the-art baselines.
Next, through systematic ablation, we investigate the catastrophic forgetting phenomenon and validate the necessity of our two-stage knowledge-insulation training strategy and the Semantic-Visual Alignment Adapter.
Furthermore, we design experiments to specifically evaluate the cognitive ability of our framework, including a dedicated Dialogue Success Rate metric.
Finally, we test the robustness of our framework in out-of-distribution environments,~\ie, under low lighting and distractors.

\subsection{Experimental Setup}
\textbf{Robot.}
We perform real-world experiments using an xArm 7~\cite{xarm7},
which has 7 DoFs and a 1-DoF gripper.
We control the xArm using the official xArm API~\cite{xarmpythonsdk}.
We use two Intel RealSense D435 cameras for the wrist-mounted and third-person viewpoints, respectively.

\textbf{Tasks.}
We design a set of real-world tasks to evaluate our framework.
These tasks strictly follow the definition in~\Cref{subsec:task-define}.
In each task, the model must first infer the correct instruction from the ambiguous one and then generate low-level actions accordingly.
We created 11 specific tasks, which fall into 4 general types:
Pick-and-Place, Pouring, Stacking, and Insertion.

\begin{itemize}
  \item Pick-and-Place: Put the fruit on the plate $\rightarrow$ Put the \texttt{Object} on the plate, where \texttt{Object} $\in$ $\{\text{Apple}, \text{Peach}, \text{Orange}\}$ (3 tasks).
  \item Pouring: Pour the water from the cup onto the plate $\rightarrow$ Pour the water from the \texttt{Color} cup onto the plate, where \texttt{Color} $\in$ $\{\text{Red}, \text{Green}, \text{White}\}$ (3 tasks).
  \item Stacking: Stack the blocks together $\rightarrow$ Stack the \texttt{Color1} block on top of the \texttt{Color2} block, where (\texttt{Color1}, \texttt{Color2}) $\in$ $\{\left(\text{Blue}, \text{Red}\right), \left(\text{Red}, \text{Blue}\right)\}$ (2 tasks).
  \item Insertion: Put the cube above the pentagon-shaped slot and drop it into the box $\rightarrow$ Put the \texttt{Color} cube above the pentagon-shaped slot and drop it into the box, where \texttt{Color} $\in$ $\{\text{Red}, \text{Green}, \text{Blue}\}$ (3 tasks).
\end{itemize}

These tasks span progressive manipulation difficulties. 
Pick-and-Place involves basic grasping and repositioning, whereas Pouring adds controlled wrist rotation and tilt. 
Stacking demands precise vertical alignment. 
Insertion imposes the strictest spatial constraints with a centimeter-level tolerance; the cube must be accurately centered above the slot to prevent rim collisions and task failure. 
Consequently, this demands fine-grained visuomotor control significantly beyond simple grasping. 
\Cref{fig:task} illustrates the tasks, workspaces, and robot setup (left to right).

\textbf{Training Data.}
For stage 1,
we use Qwen3-235B-A22B~\cite{qwen3} to synthesize 1,150 multi-turn dialogues based on 410 initial trajectory frames (trajectories collected for stage 2). Covering all 4 task types, each dialogue comprises an ambiguous instruction, 1-3 clarification QA rounds, and a resolved instruction. All data is augmented for linguistic diversity and manually verified to ensure quality.
For stage 2,
we manually collect 30-50 demonstrations for each of the 11 tasks through teleoperation~\cite{openteach} using a Meta Quest 3, resulting in 410 expert demonstrations in total.
Each demonstration is recorded at 30Hz and comprises gripper poses, states, and synchronized images from both cameras.

\textbf{Implementation Details.}
We instantiate the Cognitive Planner with Qwen2-VL-2B-Instruct~\cite{qwen2-vl} and the Motor Executor with ScaleDP-Huge~\cite{scaledp}.
The Semantic-Visual Alignment Adapter connects the two modules via a learned affine modulation mechanism: two lightweight MLPs map the observation embedding extracted by the Cognitive Planner to channel-wise scale and shift parameters, which are then applied to the corresponding prompt embedding to produce a visually-grounded condition for the Motor Executor.
We set a chunk size of 50 timesteps for our framework.

\subsection{Baselines}
We compare our framework against five baselines spanning different architectural paradigms:

\textbf{\(\boldsymbol{\pi_0}\)}~\cite{pi-0}.
A VLA with a parallel flow matching action expert.

\textbf{\(\boldsymbol{\pi_0}\)-FAST}~\cite{pi0-fast}.
An autoregressive VLA that tokenizes continuous actions.

\textbf{OpenVLA-OFT}~\cite{openvla-oft}.
A VLA with a hierarchical diffusion action expert.

\textbf{LLM-Primitives}.
A modular approach representing the ASK-TO-ACT~\cite{ask-to-act} paradigm. It first uses Qwen2-VL-72B~\cite{qwen2-vl} as a planner to conduct multi-turn dialogue for disambiguation and generate high-level action commands (\eg, ``pick\_up(blue\_block)''). These high-level actions are physically executed by a deterministic pipeline: GroundingDINO~\cite{groundingdino} and SAM~\cite{sam} perform open-vocabulary detection and segmentation to obtain 3D object poses from depth, which are then converted into joint-space trajectories via inverse kinematics and executed open-loop.

\textbf{LLM-Policy}.
A modular, non-end-to-end system that separates disambiguation from execution. It employs Qwen2-VL-72B~\cite{qwen2-vl} as a powerful planner to handle dialogue and disambiguation, reformulating the user's intent into a precise text instruction.
\input{figure/main_results}
This processed instruction is then fed to a $\pi_0$ model (fine-tuned on our data) which serves as the low-level policy for action generation. Unlike our framework, the high-level planner and low-level policy are decoupled and optimized independently.

To establish a comprehensive comparison, we delineate these baselines by their input conditions.
Specifically, \(\pi_0\), \(\pi_0\)-FAST, and OpenVLA-OFT receive unambiguous instructions during evaluation, granting them an informational advantage. Conversely, LLM-Primitives and LLM-Policy face the same ambiguous instructions as our framework, requiring active dialogue resolution before physical execution.

\textbf{Metrics.}
We evaluate our framework using two complementary metrics.
Task Success Rate (SR) measures the fraction of trials in which the robot completes the full pipeline successfully; a trial is counted only if the agent identifies the correct target through dialogue and executes the manipulation to completion.
Disambiguation Success Rate (DSR) measures the fraction of trials in which the dialogue phase correctly resolves the ambiguity in the user's instruction:
\(\mathrm{DSR} = \frac{1}{N}\sum_{i=1}^{N} \mathbf{1}\!\left(\hat{o}^{(i)} = o^{(i)}\right)\)
where $o^{(i)}$ and $\hat{o}^{(i)}$ denote the ground-truth and identified target for trial $i$, respectively; for absent-target trials, a correct refusal is also counted as a match. This metric isolates disambiguation ability from visuomotor control.

\subsection{Real-world Evaluation}

\textbf{Comparison with Baselines.}
We evaluate the \framework{} framework against five baselines across 11 real-world tasks. Each task comprises 20 trials with randomized initial object placements. To systematically validate our claims, we categorize the baselines into three distinct paradigms: (1) standard VLAs lacking dialogue capabilities (\(\pi_0\), \(\pi_0\)-FAST, OpenVLA-OFT), (2) a modular ``ASK-TO-ACT'' approach relying on high-level primitives (LLM-Primitives), and (3) a decoupled, heavy-weight pipeline combining a large language planner with a VLA policy (LLM-Policy). Results are summarized in~\Cref{fig:main-results}.

We first compare our method against the standard VLAs (\(\pi_0\), \(\pi_0\)-FAST, and OpenVLA-OFT). To isolate their manipulation capabilities, these baselines are provided with privileged, ground-truth instructions, thereby bypassing the ambiguity problem. Among them, \(\pi_0\) is the strongest, achieving a 79.1\% success rate. Remarkably, \framework{} delivers comparable performance (82.7\%) despite operating under a strictly more challenging setting where it must autonomously resolve vague user commands through multi-turn dialogue.
Notably, providing our framework directly with ground-truth instructions yields a nearly identical success rate, confirming that our disambiguation process introduces negligible performance overhead.
Furthermore, our framework substantially outperforms \(\pi_0\)-FAST (39.1\%) and OpenVLA-OFT (20.5\%).
The performance degradation in \(\pi_0\)-FAST stems from its autoregressive lossy tokenization, which sacrifices the fine-grained action fidelity essential for precise manipulation. Meanwhile, OpenVLA-OFT suffers from condition collapse (\ie, generating indistinguishable embeddings for semantically distinct instructions given the same observation, as shown in \Cref{tab:cosine-similarity}).
In contrast, our Semantic-Visual Alignment Adapter dynamically modulates instruction features via observation-conditioned transformations, ensuring robust and distinct guidance for the Motor Executor.

Furthermore, we evaluate LLM-Primitives, which represent the ``ASK-TO-ACT'' paradigm that delegates physical manipulation to high-level action primitives. Operating under the same ambiguous conditions as our framework, LLM-Primitives achieves a mere 8.6\% success rate. This failure highlights the intrinsic flaws of relying on high-level action primitives in the physical world. Inaccuracies in open-vocabulary pose estimation inevitably propagate through the deterministic trajectory planner, leading to cascading errors. This is particularly evident in the high-precision Insertion task, where LLM-Primitives completely fails (0.0\%), as even minor pose deviations cause the object to miss the narrow slot. Conversely, \framework{} utilizes closed-loop, end-to-end visuomotor control, allowing it to continuously correct spatial alignments from raw visual streams, yielding a 50.0\% success rate on the same demanding task.

Finally, we compare against LLM-Policy, a highly competitive baseline that attempts to solve the collaboration problem by coupling a massive 72B-parameter planner (Qwen2-VL-72B) for disambiguation with a fine-tuned \(\pi_0\) for execution. While this decoupled approach is highly effective (78.2\%), our integrated framework matches its performance (82.7\%) while utilizing a Cognitive Planner with merely 2B parameters.
This represents a \(36\times\) reduction in cognitive parameter count. The results demonstrate that simply stitching a powerful LLM to a VLA is computationally heavy and sub-optimal. By unifying cognitive capabilities and visuomotor control through our adapter and training strategy, \framework{} achieves state-of-the-art collaborative manipulation at a fraction of the computational cost.

\input{table/cosine_similarity}

\input{figure/knowledge_insulation}
\textbf{Impact of Training Strategies.}
We systematically ablate the training configurations in Stage 2 using the Pick-and-Place task to investigate the underlying learning dynamics.
First, we evaluate the necessity of the Semantic-Visual Alignment Adapter, designed to bridge the representation gap between the planner and the executor. As depicted in~\Cref{fig:knowledge-insulation}, directly coupling the frozen planner with the trainable executor without an adapter results in total task failure (column 5). This confirms that a cross-modal interface is strictly required to translate semantic intents into executable visual-motor features. 
However, simply introducing a naive bridge is insufficient. Replacing our adapter with a mean-pooling layer still yields a 0\% success rate across all tasks (column 4). This naive compression causes the same condition collapse as in~\Cref{tab:cosine-similarity}: given the same observation, cosine similarities between distinct instructions exceed 95\%, rendering tasks indistinguishable to the executor. This underscores the necessity of our observation-conditioned affine modulation to preserve fine-grained semantic guidance (column 3).
Furthermore, we observe that successfully bridging this gap requires a rigorous training paradigm. We investigate the impact of unfreezing different components during the visuomotor learning phase. When the planner undergoes fine-tuning in Stage 2, we observe severe catastrophic forgetting (column 1\&2). The active disambiguation capabilities cultivated in Stage 1 are completely erased, resulting in an inability to resolve user intent and leading to complete task failure.
Optimal performance (93\%) is achieved exclusively when employing our two-stage knowledge-insulation training strategy: strictly freezing the planner, while exclusively optimizing the adapter and the executor. This confirms that our design effectively aligns cognitive capabilities with visuomotor control without eroding the planner's previously acquired knowledge.

\input{figure/dsr}
\textbf{Disambiguation Ability.}
Figure~\ref{fig:dsr} illustrates the Disambiguation Success Rate (DSR) alongside inference latency, measured as the average end-to-end generation time per query, across different scales of the Qwen2-VL model (2B, 7B, 72B).
We evaluate performance in two distinct scenarios:
Present, where the target object identified through dialogue exists in the scene; and
Absent, where the target is missing, requiring the model to correctly refuse execution.
We compare our approach against two baselines:
baseline prompting, where the Qwen2-VL model receives only the raw ambiguous instruction; and
structured prompting, which explicitly instructs the Qwen2-VL model to detect ambiguity and ask clarifying questions.
Note that, similar to the baseline prompting, our planner operates directly on raw instructions without requiring prompt engineering.
As shown in Figure~\ref{fig:dsr}, our 2B planner achieves the highest DSR across both \textit{Present} and \textit{Absent} settings.
In the \textit{Present} setting, our model outperforms the structured 2B and 7B models and matches the 72B model.
In the \textit{Absent} setting, where the model must realize the target is missing, our method consistently issues correct refusals, significantly surpassing the structured 2B model.
Overall, our 2B planner matches the performance of the structured 72B model while using $36\times$ fewer parameters.
In contrast, under the comparable \textit{baseline prompting} setting, all Qwen2-VL models struggle significantly, exhibiting notably deficient DSR.
Furthermore, regarding computational efficiency, our approach demonstrates the lowest inference latency, matching the baseline Qwen2-VL-2B model.
We attribute this superior DSR performance to our two-stage knowledge-insulation training strategy.
This strategy effectively preserves the disambiguation knowledge acquired during training, preventing catastrophic forgetting.
Consequently, our method internalizes the ability to disambiguate and refuse impossible tasks through dialogue.
By achieving the highest DSR concurrently with the lowest inference latency, it eliminates the need for prompt engineering or model scaling.    

\input{figure/ood_results}
\textbf{Low-Light Conditions.}
To assess visual robustness, we reduce illumination by 50\% in the evaluation of the Stacking task. 
It can be observed in~\Cref{fig:ood-results} that the performance of both the~\framework{} framework and LLM-Policy drops, but the impact is much smaller for our framework. Its success rate decreases only slightly, from 90.0\% to 80.0\%. In contrast, LLM-Policy's success rate drops sharply from 92.5\% to 22.5\%.
This could be due to the fact that our framework keeps the VLM's vision encoder frozen, preserving its general robustness to visual distribution shifts.
The baseline LLM-Policy, however, requires fine-tuning its entire policy ($\pi_0$) to work on our tasks. This makes it overfit to the training conditions and less robust to visual changes.
The result suggests that our framework is robust under low-light conditions.

\textbf{Distractors.}
We test the robustness of the~\framework{} framework in the presence of visual distractors.
Specifically, we place both an apple and a pomegranate on the table, which appear strikingly similar under our cameras.
As shown in~\Cref{fig:ood-results}, our framework shows a more robust ability with a distractor present.
It maintains a high success rate of 80.0\%, while LLM-Policy's performance falls to 65.0\%.
The results indicate our framework can better handle confusing objects, a common challenge in real-world scenarios.

%% file: figure/task.tex
\begin{figure*}[ht]
\centering
\includegraphics[width=.99\linewidth]{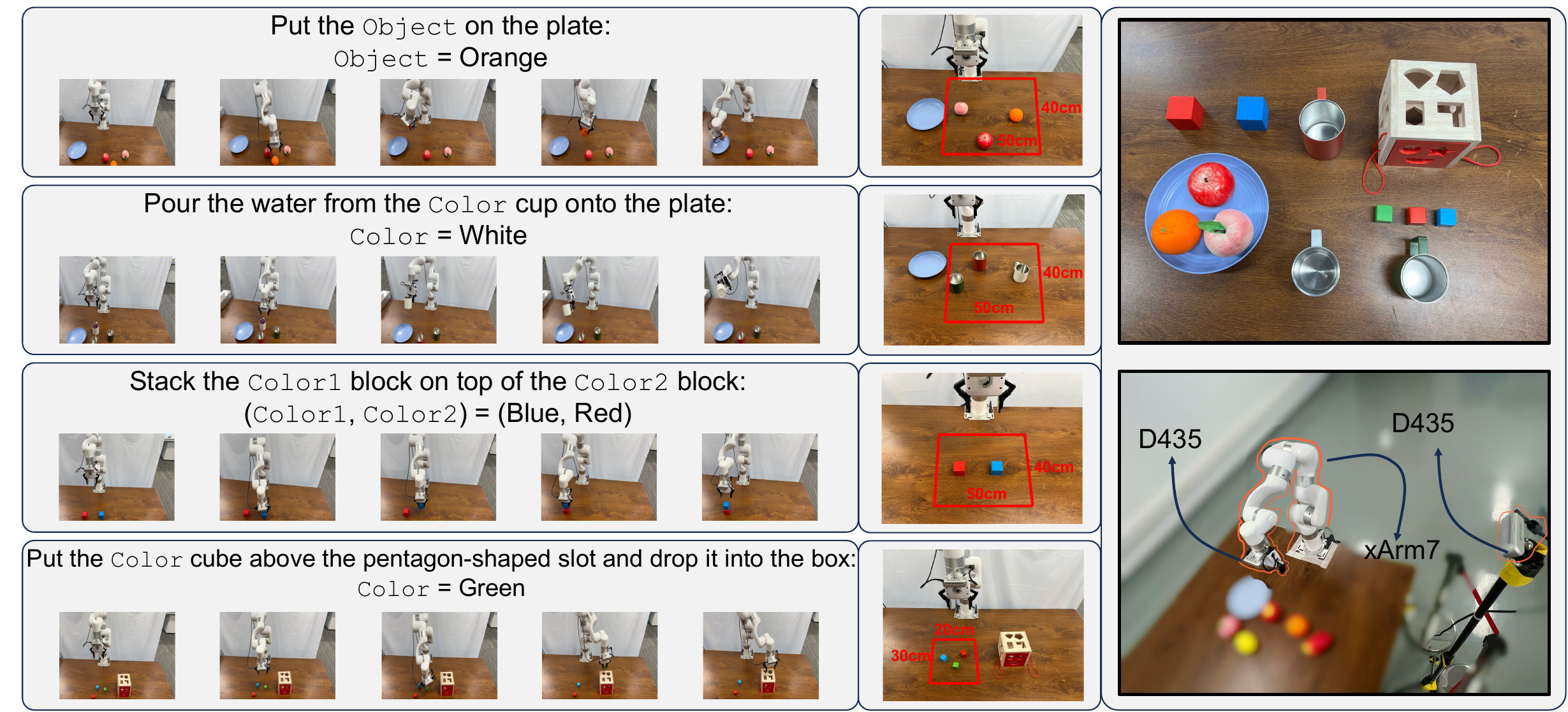} 
  \caption{
  \textbf{Experimental setup.}
  \textbf{Left:} Execution sequences for the four task categories: Pick-and-Place, Pouring, Stacking, and insertion.
  \textbf{Middle:} Red bounding boxes indicating randomized object initialization zones.
  \textbf{Right:} The object inventory (top) and the xArm7 robot equipped with two Intel RealSense D435s (bottom).
  }

\label{fig:task}
\end{figure*}

%% file: figure/main_results.tex
\begin{figure*}[ht]
\centering
\includegraphics[width=.85\linewidth]{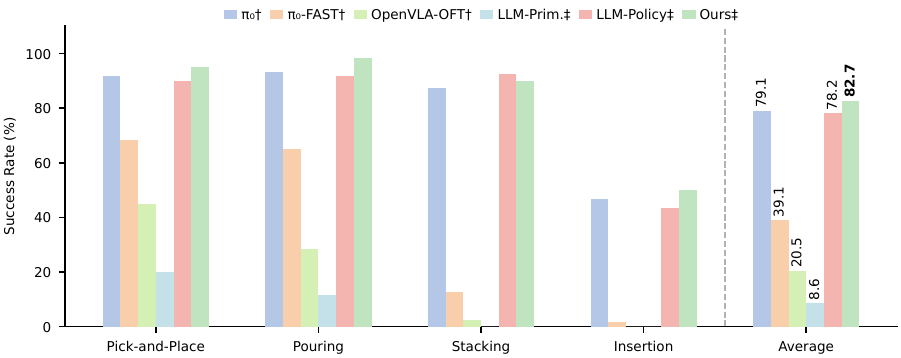} 
  \caption{
  \textbf{Real-world evaluation results across 11 tasks.}
  $\dagger$: receives correct instructions directly.
  $\ddagger$: receives ambiguous instructions and must resolve ambiguity through dialogue.
  }
\label{fig:main-results}
\end{figure*}

%% file: table/cosine_similarity.tex
\begin{table}[htb]
\centering
\caption{\textbf{The similarity matrix.}
}
\begin{tabular}{c!{\vrule width \lightrulewidth}c!{\vrule width \lightrulewidth}c!{\vrule width \lightrulewidth}c} 
\toprule
\multicolumn{1}{l!{\vrule width \lightrulewidth}}{}
& Apple & Orange & Peach\\ 
\midrule
Apple & 1 & 98.44\% & 98.83\% \\
\midrule
Orange & 98.44\% & 1 & 99.22\% \\
\midrule
Peach & 98.83\% & 99.22\% & 1 \\
\bottomrule
\end{tabular}

\label{tab:cosine-similarity}
\end{table}

%% file: figure/knowledge_insulation.tex
\begin{figure}[ht]
\centering
\includegraphics[width=.85\linewidth]{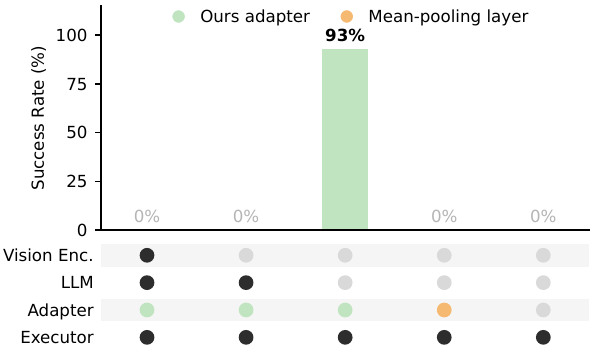} 
\caption{
\textbf{Performance comparison of different training strategies in stage 2.}
}
\label{fig:knowledge-insulation}
\end{figure}

%% file: figure/dsr.tex
\begin{figure}[ht]
\centering
\includegraphics[width=.99\linewidth]{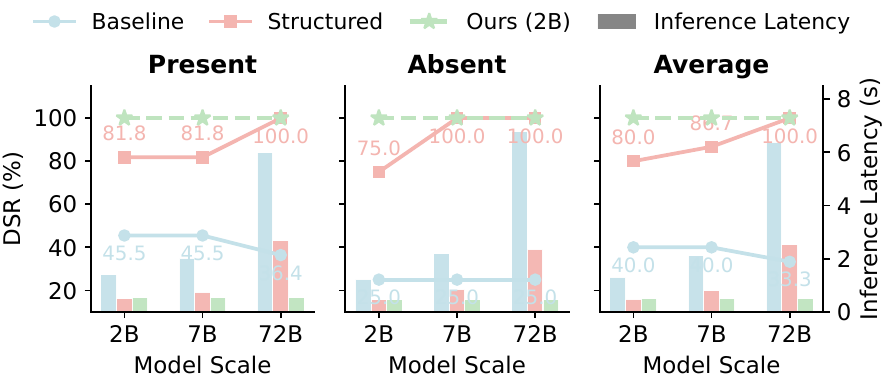} 
\caption{\textbf{Disambiguation Success Rate (lines, left axis) and inference latency (bars, right axis) across model scales.}}

\label{fig:dsr}
\end{figure}

%% file: figure/ood_results.tex
\begin{figure}[ht]
\centering
\includegraphics[width=.85\linewidth]{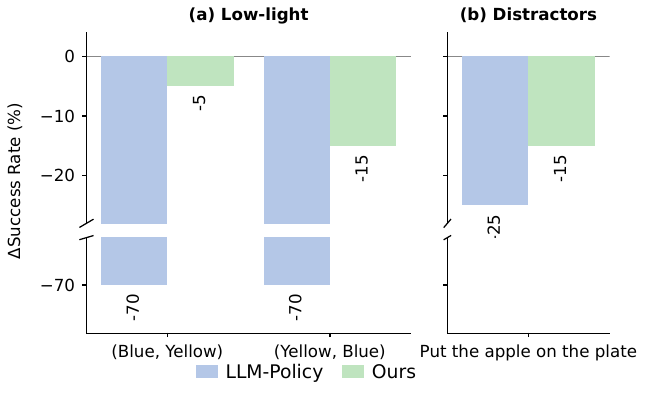} 
\vspace{-0.3em}
\caption{
\textbf{OOD robustness evaluation.}
Bars show success rate (\%) changes relative to IID conditions.
}
\label{fig:ood-results}
\end{figure}

%% file: tex/5_conclusion.tex
We present \framework{}, a unified end-to-end framework integrating multi-turn disambiguation dialogue with low-level visuomotor control.
Unlike conventional ``Listen-and-Act'' systems, our approach resolves ambiguity through active collaboration before executing precise physical manipulation.
Specifically,~\framework{} synergizes a VLM-based Cognitive Planner with a Diffusion-based Motor Executor. 
To bridge the inherent representation gap between high-level cognitive capabilities and low-level visuomotor control, we introduce the Semantic-Visual Alignment Adapter, which ensures visually-grounded conditioning for the executor.
Furthermore, to prevent the severe catastrophic forgetting of cognitive capabilities during downstream visuomotor fine-tuning, we employ a two-stage knowledge-insulation training strategy. This strategy strictly freezes the planner, effectively decoupling disambiguation from execution while preserving previously acquired knowledge. 
During deployment, a Disambiguation Gate seamlessly orchestrates the transition between disambiguation and execution modes. 
Empirical results across real-world tasks demonstrate that~\framework{} significantly outperforms state-of-the-art baselines, offering an informative paradigm for collaborative embodied agents.

\textbf{Future Work}
Our future research will focus on enhancing personalization and interactivity by integrating long-term memory to retain user preferences and proactively anticipate intent. 
Additionally, transitioning to natural voice communication will enable more intuitive dialogue for dynamic real-world environments.